\documentclass[sigconf]{acmart}

\usepackage{amsmath}
\usepackage{multirow}
\usepackage{booktabs}
\usepackage{subcaption}

\usepackage{graphicx}

\usepackage[utf8]{inputenc}
\usepackage{amsthm}
\usepackage{amssymb}
\usepackage{algorithm}
\usepackage{comment}
\usepackage{graphicx}
\usepackage[skip=0pt,textfont=bf,labelfont=bf]{caption}
\usepackage{lipsum}
\usepackage{array}

\setcopyright{acmlicensed}
\acmDOI{xxxxx}

\acmConference[ SIGIR'18 Workshop, EARS'18]{SIGIR}{July'18}{Michigan, USA}
\acmYear{2018}
\copyrightyear{2018}

\acmArticle{x}
\acmPrice{yy}
\setcopyright{rightsretained}
\acmConference[EARS'18]{SIGIR 2018 Workshop on ExplainAble Recommendation and
Search}{July 12, 2018}{Ann Arbor, Michigan, USA}
\begin{document}
\title{Layer-wise Relevance Propagation for Explainable Recommendations}

\author{Homanga Bharadhwaj}
\affiliation{%
  \institution{Department of Computer Science and Engineering\\ Indian Institute of Technology Kanpur, India}
}
\email{}




\begin{abstract}
    In this paper, we tackle the problem of explanations in a deep-learning based model for recommendations by leveraging the technique of layer-wise relevance propagation. We use a Deep Convolutional Neural Network to extract relevant features from the input images before identifying similarity between the images in feature space. Relationships between the images are identified by the model and layer-wise relevance propagation is used to infer pixel-level details of the images that may have significantly informed the model's choice.  We evaluate our method on an Amazon products dataset and demonstrate the efficacy of our approach.   
\end{abstract}
\maketitle




\section{Introduction}

Explainability in recommendation systems (RS) has been a topic of great interest for a long time now~\cite{ex1,ex2,ex3}. Although the topic of explainability is fairly ubiquitous and is of general interest in the Artificial Intelligence (AI) community~\cite{xai1,xai2,xaideep}, it is especially important for recommendation to persuade users into accepting the recommended items. Since the aim of most recommendation systems is greater commercial viability, an ill-informed recommendation could lead to a long-term loss of users' trust. Hence, the stakes in modern recommendation systems are quite high with robust and interpretable recommendations being the need of the hour.

There have been recent works leveraging images for recommendation~\cite{shankar2017deep,mcauley2015image}. Deep Convolutional Neural Networks (DCNNs) have especially been of great utility in this regard and have proved to be more robust and reliable. For recommendations, ideas for using pixel-level attention have been explored in a few recent papers~\cite{zhou2017deep,chen2017attentive}. Explainability for deep learning based models is extremely challenging and is an active topic of current research~\cite{lapuschkin2016analyzing,samek2017evaluating,bach2015pixel}. However, there have been recent attempts to explain the predictions of classifiers in terms of features of the input. Layer-wise relevance propagation seeks to explain a classifier's decisions by decomposition~\cite{lapuschkin2016analyzing,bach2015pixel}. It redistributes the final prediction output backwards in the network so as to eventually assign relevant scores to each input variable (in the case of images, input variables are the image pixels). 

In this work, we extend the framework of Layer-wise relevance propagation to the RS setting. The task of inferring items (images) relevant to a query item (image), such that the inferred items complement / substitute the query item is an important task in online shopping platforms like Amazon. We adopt a similar problem statement as McAuley et al.~\cite{mcauley2015image} and develop a DCNN based architecture similar to that proposed in the paper. The novelty in our approach is that we train our model end-to-end and incorporate layer-wise relevance propagation in this system so that features of the recommended images which were most central to the recommendation task are identified.  Through detailed simulation results on the dataset used by McAuley et al.~\cite{mcauley2015image}, we show that including the layer-wise relevance feedback in our model does produce effective explanations by employing the method of perturbations~\cite{lapuschkin2016analyzing}. We elaborate in the future works that the information of relevance of pixels in the explanations can be further integrated into the training method so as to enable a more robust identification scheme.

\section{The Methodology}

Our basic aim is to develop a method for inferring preference of users for the visual appearance of one object given a query image. To begin, we project every image in the dataset into a $K$-dimensional feature vector with the help of a DCNN. An important distinction between~\cite{mcauley2015image} and our method is that we do not use a pre-trained network. This is because, we intend to integrate layer-wise relevance propagation to improve the predictions of the model. Our approach is independent of the specific architecture of the DCNN used. The second last (fully-connected) layer results in the desired features of each input image. Similar to~\cite{mcauley2015image}, we learn a distance transform between feature vectors (that correspond to images) and use the distances to infer the probability of relevance of one image to another. The distance metric is personalized for each user since we are interested in evaluating personalized recommendations. 

\subsection{Personalized Recommendations}

Let $\mathbf{x_i}$ and $\mathbf{x_j}$ denote the feature vectors of the $i^{th}$ item and the $j^{th}$ item respectively. Then, we define the distance function $d_u(\mathbf{x_i},\mathbf{x_j})$ to represent the distance between the $i^{th}$ item and the $j^{th}$ item according to the user $u$ as 
\begin{equation}
    d_u(\mathbf{x_i},\mathbf{x_j}) = (\mathbf{x_i} - \mathbf{x_j}) \mathbf{M}^{(u)} (\mathbf{x_i} - \mathbf{x_j})^T
\end{equation}
Here, $\mathbf{M}^{(u)}$ is a personalized weighing of the importance of each feature for user $u$. $\mathbf{M}$ is a square matrix of $D$ dimensions. This distance metric is used to infer whether objects $i$ and $j$ are related. Let $R^u$ be the set of relationships such that $r_{ij}\in R^u$ relates objects $i$ and $j$ for user $u$. Section 3.1 describes the various relationships present in the dataset. We use a shifted sigmoid function of the distance $d_u(\mathbf{x_i},\mathbf{x_j})$ to infer $P_u(r_{i,j}\in R )$ i.e. the probability of objects $i$ and $j$ being related.    

\begin{equation}
    P_u(r_{i,j}\in R) = \frac{1}{1 + e^{d_u(\mathbf{x_i},\mathbf{x_j})-q}}  
\end{equation}
Here, $q$ is tuned during training so as to maximize prediction accuracy. 

\subsection{Training}

Our complete model consists of a DCNN architecture to obtain relevant features from the image followed by a layer for computing the distances between the different feature vectors obtained from the DCNN. A VGG-16 model is used and the last fully-connected layer serves as the feature extractor. For training, we have an objective similar to that proposed in~\cite{mcauley2015image}, but a major difference is that we propagate gradients back through the DCNN as well and we have explicitly considered personalization. We first randomly sample a negative set $S = \{ r_{ij} | r_{ij} \notin R \}$ by maintaining $|R| = |S|$. This is a standard practice in training, wherein the probability of positive examples in maximized while minimizing the probability of neagtive examples.
\begin{align*}
    L(\mathbf{M},\mathbf{\Theta},q|R,S) = \sum_{u} (\sum_{r_{ij}\in R}-\log(1 + e^{d_u(\mathbf{x_i},\mathbf{x_j})-q} ) + \\ \sum_{r_{ij}\in S}(1 + \log(1 + e^{d_u(\mathbf{x_i},\mathbf{x_j})-q} ))  )
\end{align*}
Here, $\mathbf{\Theta}$ refers to all the parameters of the DCNN from the first layer to the feature extraction (pre-final) layer. We train our model end-to-end by performing standard backpropagation over the entire network by taking gradients of $L$ wrt the relevant parameters. We use ADAM optimizer for gradient descent~\cite{kingma2014adam}. 

\subsection{Layer-wise relevance propagation}

This is a general technique that can be applied to most NN architectures~\cite{arras2017explaining,lapuschkin2016analyzing,bach2015pixel}. Here, we apply it to our entire DCNN framework. The model infers relationships between pairs of input images and eventually also recommends items based on a query image ( Section 3.4). Let $g(\mathbf{x})$ be the model's prediction. This prediction is redistributed to each input pixel and a relevance score $R_i$ is assigned to each input pixel $i$. The central idea of this relevance propagation is relevance conservation i.e. $ \sum_i R_i^{(1)} =  .... = \sum_j R_j^{(l)} = .... = g(\mathbf{x})$, where $l$ denotes a generic layer of the network. This implies that the total relevance is conserved at each layer.

This essentially means that at each layer of the DCNN, the total relevance which equals the prediction $g(\mathbf{x})$ is conserved. The relevance score of each input variable determines how much this variable has contributed to the prediction. Consider a neuron in our DCNN. It maps a set of inputs $x_i$ to an output $x_j$ through a combination of weights $w_{ij}$ and an activation function, let us call it $h(.)$. So,
\begin{equation}
    x_j = h \left(\sum_i w_{ij}x_i \right)
\end{equation}

The Relevance assignment mechanism works by computing a relevance $R_i$ for neuron $x_i$ (input) when all the relevances $R_j$ of outputs $x_j$ are given. We use the following formula for this propagation, although many variants exist~\cite{bach2015pixel}.

\begin{equation}
    R_i^{(l)} = \sum_j \frac{x_i^{(l)} w_{ij}^{(l,l+1)}}{\sum_k x_k^{(l)} w_{kj}^{(l,l+1)} + \epsilon\times sign(\sum_k x_k^{(l)} w_{kj}^{(l,l+1)})} R_j^{(l+1)}
\end{equation}

The output of layer-wise relevance propagation is essentially a heatmap of the image. Evaluating how good (informative) is the heatmap is tricky. The method of perturbations was proposed in~\cite{samek2017evaluating}. The key idea is that if one perturbs (changes the value of) the highly important input variables as predicted by the model, the decline in prediction score should be steeper than if other less important variables are perturbed. Using an iterative scheme to perturb input variables, we have an objective measure of explanation quality - larger decline in prediction accuracy corresponds to a more successful explanation scheme.

\section{Experiments}

\subsection{Setup}

We used Tensorflow r1.4~\cite{abadi2016tensorflow} and Python3 for all relevant programming. The dataset used is from~\cite{mcauley2015image}. The details of the dataset including the various sub-categories are illustrated in detail in this paper~\cite{mcauley2015image}. The open source code available at~\cite{lapuschkin2016analyzing} was used (after significant modifications) for incorporating the layer-wise relevance propagation in our architecture. There are four types of relationships between items (say A and B) in the dataset, namely  1) People viewed A and also viewed B, 2) People viewed A and ended up buying B, 3) People bought A and also later bought B, 4) People bought A and B together. The categories 1) and 2) are broadly categorized as `substitutes' as the products A and B have some notion of being substitutable while 3) and 4) are broadly categorized as `complements'. Section 1.1 of~\cite{mcauley2015image} elaborates more on the dataset. We perform a two-fold analysis, namely generating recommendations and inferring explanations. For effective comparison and benchmarking, all our experiments follow the same protocol, including the same train-test split of data outlined in section 4.1 of~\cite{mcauley2015image}.

\subsection{Results}
Our recommendation method differs from~\cite{mcauley2015image} with respect to the architecture of the DCNN and the fact that we did not use a pre-trained DCNN model as a feature extractor. By incorporating an end-to-end training mechanism, we obtained slightly better results that that reported in~\cite{mcauley2015image}. We used a VGG-16 model in the initial part of our pipeline such that its pre-final fully connected layer acts as a feature extractor. For initialization, all the model weights $\mathbf{\Theta}$ are assigned as per the pre-trained model on Imagenet dataset~\cite{deng2009imagenet}. To make things clear, this network is not fixed in our pipeline, but the parameters $\mathbf{\Theta}$ are updated during training unlike~\cite{mcauley2015image} (We use the pre-trained model to only initialize the weights). Table~\ref{tb:results} shows the results of this model on the test set after end-to-end training.

\begin{table}[]
\centering
\setlength\tabcolsep{2pt}
\caption{Results ($\%$ accuracy) averaged over items in each category for each of the four relationship types. $D$ refers to the dimension of the weighing matrix defined in Section 2.1. End-to-end training leads to better results than that reported in~\cite{mcauley2015image}. }
\label{tb:results}
\begin{tabular}{cccccc}
\hline
Category                                                                                 & D & \begin{tabular}[c]{@{}c@{}}Buy after\\ viewing\end{tabular} & \begin{tabular}[c]{@{}c@{}}Also\\ viewed\end{tabular} & \begin{tabular}[c]{@{}c@{}}Also\\ bought\end{tabular} & \begin{tabular}[c]{@{}c@{}}Bought\\ together\end{tabular} \\ \hline
\multirow{3}{*}{Books}                                                                   & 10         & 70.3                                                        & 68.9                                                  & 69.1                                                  & 68.8                                                      \\
                                                                                         & 100        & 72.6                                                        & 70.4                                                  & 72.2                                                  & 68.3                                                      \\
                                                                                         & 500        & 72.4                                                        & 70.6                                                  & 72.3                                                  & 68.3                                                      \\ \hline
\multirow{3}{*}{\begin{tabular}[c]{@{}c@{}}Cell Phones\\ and Accessories\end{tabular}}   & 10         & 84.4                                                        & 78.8                                                  & 78.8                                                  & 84.2                                                      \\
                                                                                         & 100        & 86.0                                                        & 83.7                                                  & 84.1                                                  & 88.3                                                      \\
                                                                                         & 500        & 86.3                                                        & 83.5                                                  & 84.2                                                  & 88.3                                                      \\ \hline
\multirow{3}{*}{\begin{tabular}[c]{@{}c@{}}Clothing, Shoes\\ and\\ Jewelry\end{tabular}} & 10         & -                                                           & 77.3                                                  & 74.1                                                  & 78.4                                                      \\
                                                                                         & 100        & -                                                           & 87.8                                                  & 85.0                                                  & 89.8                                                      \\
                                                                                         & 500        & -                                                           & 88.1                                                  & 85.4                                                  & 89.9                                                      \\ \hline
\multirow{3}{*}{\begin{tabular}[c]{@{}c@{}}Digital \\ Music\end{tabular}}                & 10         & 68.7                                                        & 61.1                                                  & 74.6                                                  & 55.2                                                      \\
                                                                                         & 100        & 73.1                                                        & 63.9                                                  & 76.8                                                  & 61.2                                                      \\
                                                                                         & 500        & 73.3                                                        & 63.9                                                  & 76.7                                                  & 62.2                                                      \\ \hline
\multirow{3}{*}{\begin{tabular}[c]{@{}c@{}}Electronic\\ Items\end{tabular}}              & 10         & 83.9                                                        & 80.5                                                  & 78.0                                                  & 79.4                                                      \\
                                                                                         & 100        & 85.8                                                        & 84.1                                                  & 83.2                                                  & 85.1                                                      \\
                                                                                         & 500        & 86.1                                                        & 84.3                                                  & 83.2                                                  & 85.1                                                      \\ \hline
\multirow{3}{*}{\begin{tabular}[c]{@{}c@{}}Grocery and \\ Gourmet Food\end{tabular}}     & 10         & -                                                           & 78.1                                                  & 82.7                                                  & 79.6                                                      \\
                                                                                         & 100        & -                                                           & 79.2                                                  & 86.9                                                  & 86.4                                                      \\
                                                                                         & 500        & -                                                           & 83.1                                                  & 86.8                                                  & 86.3                                                      \\ \hline
\multirow{3}{*}{\begin{tabular}[c]{@{}c@{}}Home and \\ Kitchen\end{tabular}}             & 10         & 78.5                                                        & 81.3                                                  & 79.7                                                  & 79.5                                                      \\
                                                                                         & 100        & 82.4                                                        & 83.8                                                  & 80.8                                                  & 84.2                                                      \\
                                                                                         & 500        & 82.6                                                        & 84.1                                                  & 81.0                                                  & 84.7                                                      \\ \hline
\multirow{3}{*}{\begin{tabular}[c]{@{}c@{}}Movies \\ and TV\end{tabular}}                & 10         & 72.4                                                        & 69.9                                                  & 73.2                                                  & 68.4                                                      \\
                                                                                         & 100        & 73.8                                                        & 69.8                                                  & 79.4                                                  & 69.6                                                      \\
                                                                                         & 500        & 73.9                                                        & 70.3                                                  & 79.2                                                  & 69.8                                                      \\ \hline
\multirow{3}{*}{\begin{tabular}[c]{@{}c@{}}Musical \\ Instruments\end{tabular}}          & 10         & 84.6                                                        & 87.9                                                  & 85.7                                                  & 82.2                                                      \\
                                                                                         & 100        & 90.3                                                        & 88.5                                                  & 85.9                                                  & 84.6                                                      \\
                                                                                         & 500        & 90.9                                                        & 88.5                                                  & 86.4                                                  & 84.7                                                      \\ \hline
\multirow{3}{*}{\begin{tabular}[c]{@{}c@{}}Office \\ Products\end{tabular}}              & 10         & 81.9                                                        & 84.6                                                  & 84.5                                                  & 78.5                                                      \\
                                                                                         & 100        & 86.8                                                        & 88.6                                                  & 87.2                                                  & 81.8                                                      \\
                                                                                         & 500        & 87.1                                                        & 88.8                                                  & 87.6                                                  & 82.3                                                      \\ \hline
\multirow{3}{*}{\begin{tabular}[c]{@{}c@{}}Toys and \\ Games\end{tabular}}               & 10         & 75.7                                                        & 79.1                                                  & 78.8                                                  & 81.1                                                      \\
                                                                                         & 100        & 77.2                                                        & 82.8                                                  & 82.4                                                  & 83.7                                                      \\
                                                                                         & 500        & 77.5                                                        & 83.0                                                  & 82.7                                                  & 83.6                                                      \\ \hline
\end{tabular}
\end{table}

\subsection{Explanations for item relationships}

This section presents our results on how well layer-wise relevance propagation is able to generate explanations. Let us first clearly define what the explanations mean in our context. For each category, say the category `also viewed', the image pixels of both the images between whom the relationship is being considered which are more relevant to the relationship prediction task are highlighted. Note that this is different from the conventional relevance prediction of image pixels for a Deep Neural Network classifier, where the image pixels of the image that are instrumental to identification of the correct object in the image ( say a 'man') are highlighted. In our context, the pixels in both the images which lead to a correct identification of their relationship are highlighted.

\begin{figure}[t]
    \centering
    \begin{subfigure}[b]{0.45\columnwidth}
        \includegraphics[width=1\columnwidth]{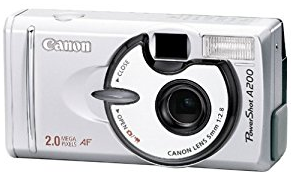}
    \end{subfigure}
    ~
    \begin{subfigure}[b]{0.5\columnwidth}
        \includegraphics[width=1\columnwidth]{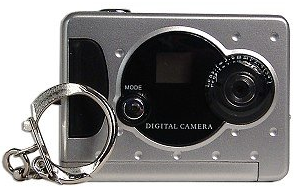}
    \end{subfigure}
    \caption{Two images of camera (in category Electronics) that are related via `also viewed' which the model predicts correctly}\label{fig:features}
\end{figure}

\begin{figure}[t]
    \centering
    \begin{subfigure}[b]{0.45\columnwidth}
        \includegraphics[width=1\columnwidth]{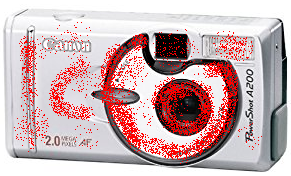}
    \end{subfigure}
    ~
    \begin{subfigure}[b]{0.5\columnwidth}
        \includegraphics[width=1\columnwidth]{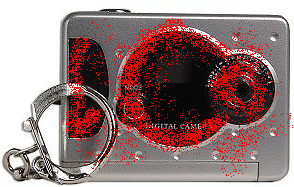}
    \end{subfigure}
    \caption{Heatmaps of the two images superimposed on the original image for clarity. The highlighted pixels are those that layer-wise relevance propagation deems most suitable for informing the model about the relationship between these two images}\label{fig:explain}
\end{figure}

\begin{figure}[t]
 \centering
        \includegraphics[width=1\columnwidth]{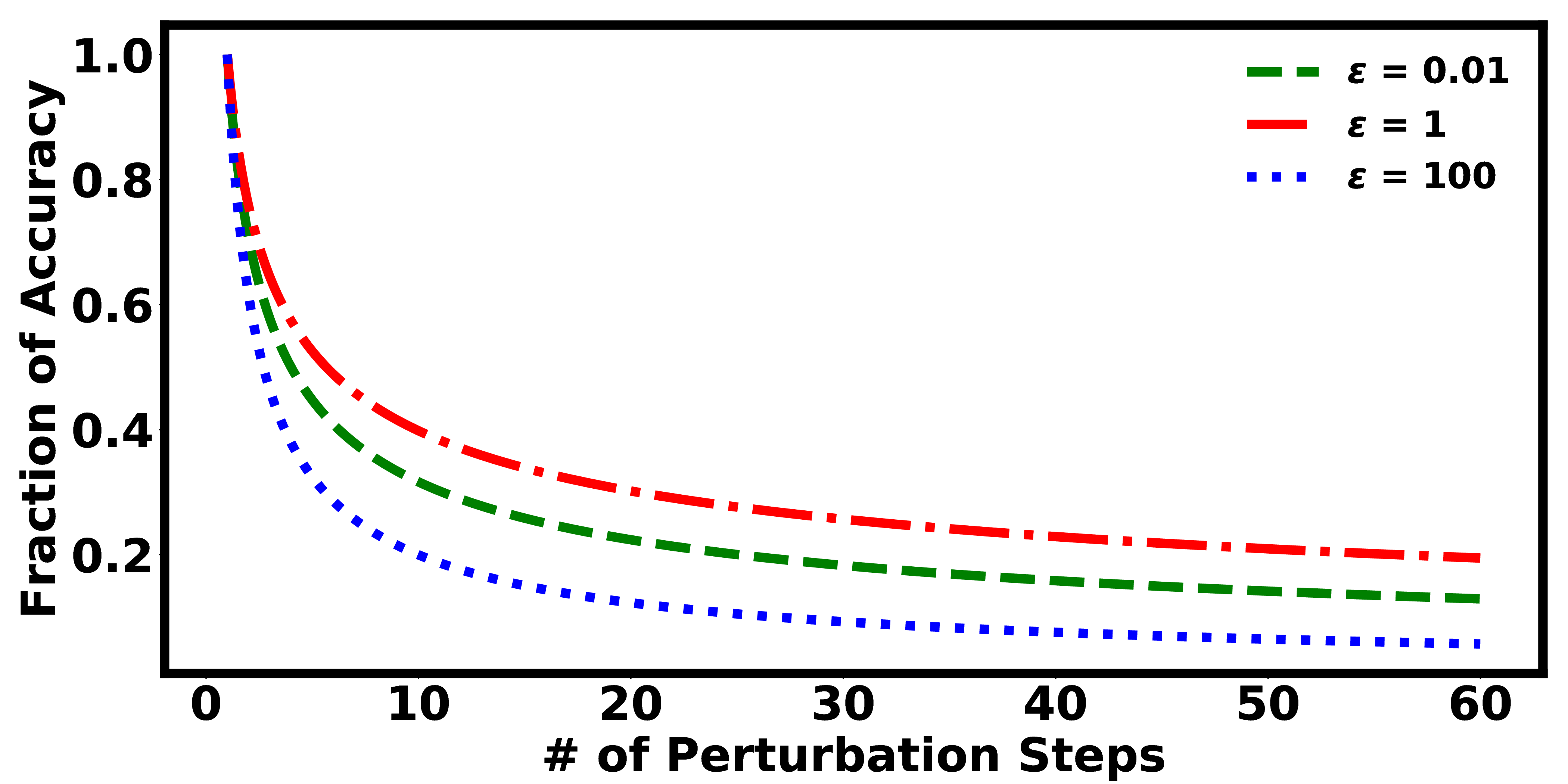}
        \caption{Layer-wise relevance propagation for different values of $\epsilon$. The decrease in accuracy averaged over the entire test-set for all categories is shown as a function of the number of perturbations. The greater the decrease in accuracy the better, because it implies that the explanations indicating relevancy of pixels is more accurate.}
        \label{fig:lrp}
    \end{figure}

To judge the efficacy of the explanations (highlighted pixels), the method of perturbations is employed~\cite{lapuschkin2016analyzing}. A random perturbation scheme is employed, in that input variables (pixels) are replaced by a random sample from a uniform distribution. We perform a category-wise analysis. For example, given an image of a book, we look at an image that is related to this image by the relation `buy after viewing' and is actually predicted correctly by the model. Now, after perturbation, we look at what relationship this image has with the original image (as predicted by the model). This is done for all the images in each category and we report the results averaged over all categories in Figure~\ref{fig:lrp}. Figures~\ref{fig:features} and~\ref{fig:explain} present anecdotal evidence of the explanations for two `camera' items.

\subsection{Recommendations and their explanations}

The pipeline described so far identifies relationships between the products but does not explicitly recommend items. In this section, we show how this model can be used for generating recommendations and also present anecdotal evidence of the process.  The recommendation task is formulated as, given a query item ( which could be a product being currently looked upon by the user, or a recently purchased) , we recommend a set of other items `to go along with it' or that the user might be interested in. Since we have already leveraged the relationships between items, this task is straightforward.

Following the approach of~\cite{mcauley2015image}, for the user $u$ given the feature vector $\mathbf{x_i}$ of the query item, the following is used to generate recommendations for each category $C$,

\begin{equation}
    \arg\max_{j\in C} P_u(r_{i,j}\in R)
\end{equation}

\begin{figure}[t]
    \centering
    \begin{subfigure}[b]{0.22\columnwidth}
        \includegraphics[width=1\columnwidth]{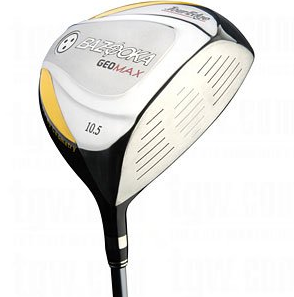}
    \end{subfigure}
    ~
    \begin{subfigure}[b]{0.22\columnwidth}
        \includegraphics[width=1\columnwidth]{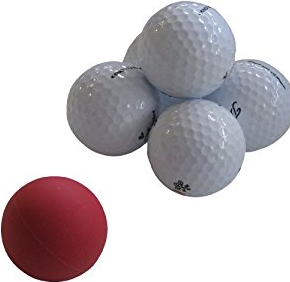}
    \end{subfigure}
    ~
    \begin{subfigure}[b]{0.22\columnwidth}
        \includegraphics[width=1\columnwidth]{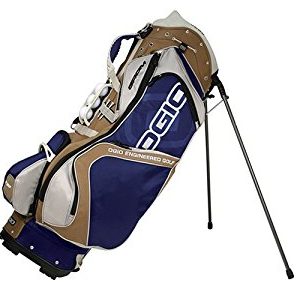}
    \end{subfigure}
    ~
    \begin{subfigure}[b]{0.22\columnwidth}
        \includegraphics[width=1\columnwidth]{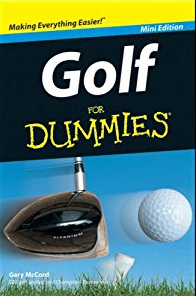}
    \end{subfigure}
    \caption{The first image on the left is the query. The remaining three are recommendations. The model correctly recommends `golf' related items given the query image. }\label{fig:rec}
\end{figure}

The authors of~\cite{mcauley2015image} do not consider personalization here, but in our approach we have used the personalized distance metric for inferring  $P_u(r_{i,j}\in R)$ as mentioned in Section 3. Figure~\ref{fig:rec} shows anecdotal evidence for this recommendation process. Without showing the long temporal history of users (which is not possible to illustrate here due to space constraint), how personalized the recommendations are cannot be ascertained. However, we can clearly see that complementary items to the query item are being recommended by the model. 

\begin{figure}[t]
    \centering
    \begin{subfigure}[b]{0.22\columnwidth}
        \includegraphics[width=1\columnwidth]{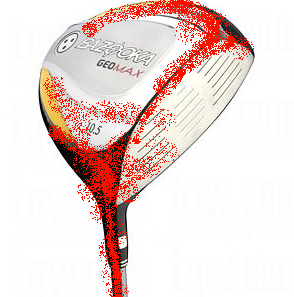}
    \end{subfigure}
    ~
    \begin{subfigure}[b]{0.22\columnwidth}
        \includegraphics[width=1\columnwidth]{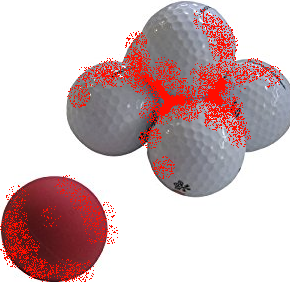}
    \end{subfigure}
    ~
    \begin{subfigure}[b]{0.22\columnwidth}
        \includegraphics[width=1\columnwidth]{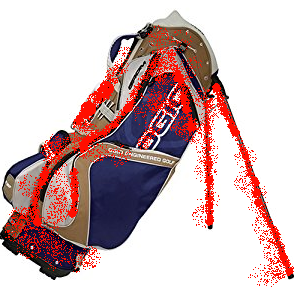}
    \end{subfigure}
    ~
    \begin{subfigure}[b]{0.22\columnwidth}
        \includegraphics[width=1\columnwidth]{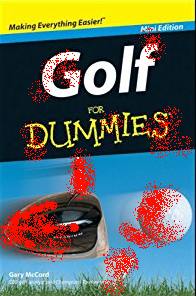}
    \end{subfigure}
    \caption{Application of layer-wise relevance propagation denotes highlighted pixels that are likely to have been most informative for the recommendation task. The heatmaps are superimposed on the original images for clarity and better comparison.  }\label{fig:recexplain}
\end{figure}

In Section 3.3, we illustrated explanations for the task of inferring correct relationships between items. In Figure~\ref{fig:recexplain} we demonstrate the application of layer-wise relevance propagation to the above mentioned recommendation task. The heatmaps do appear a little noisy, indicating less certainty about relevant pixels. It is a part of our future work to investigate better schemes for tuning the layer-wise relevance propagation. To be clear about what explanations mean in this context, the highlighted pixels are the most informative (as predicted by layer-wise relevance propagation) for inferring which items will be complementary to the query item. We restrict the number of outputs to three here, but it is a modelling choice and can be varied. 

\section{Future Works}

The major future work with respect to the method outlined in this paper is using the explanations to improve the model's predictions. Improving predictions include improving the extracted features and eventually inferring correctly the relationships between different items. These will eventually enable the model to give better recommendations. This is technically not a `future' work because we are already in the process of implementing it. We are incorporating the relevant pixels by configuring the convolutional layers to give more weight to the regions of the image where highly relevant pixels are present.

This is essentially a meta-learning process and the integration of relevant pixels in the training should be done through the validation set. The model is allowed to make predictions on the validation set, the explanations for these predictions are generated by layer-wise relevance feedback and the information from the relevant pixels are used to modify the weights in the convolutional layers of the network. So, including explanations for training essentially entails an outer training loop over the existing framework.

\section{Conclusion}
In this paper, we investigated the application of layer-wise relevance propagation for generating explanations in a deep-learning based recommendation framework. We developed an end-to-end mechanism for training the DCNN to generate relevant features and infer relationships between items. By applying layer-wise relevance propagataion to the entire framework, we showed through simulation studies that informative pixels are identified by the model. In addition, we explicitly considered the task of recommendation and generated explanations for the same. As highlighted in the Future Works, a lot remains to be done and we are hopeful that this direction of research will help in using explanations for improving modelling accuracy as well.


\bibliographystyle{abbrv}
\bibliography{sigproc}  

\begin{thebibliography}{10}

\bibitem{abadi2016tensorflow}
M.~Abadi, P.~Barham, J.~Chen, Z.~Chen, A.~Davis, J.~Dean, M.~Devin,
  S.~Ghemawat, G.~Irving, M.~Isard, et~al.
\newblock Tensorflow: A system for large-scale machine learning.
\newblock In {\em OSDI}, volume~16, pages 265--283, 2016.

\bibitem{arras2017explaining}
L.~Arras, G.~Montavon, K.-R. M{\"u}ller, and W.~Samek.
\newblock Explaining recurrent neural network predictions in sentiment
  analysis.
\newblock {\em arXiv preprint arXiv:1706.07206}, 2017.

\bibitem{bach2015pixel}
S.~Bach, A.~Binder, G.~Montavon, F.~Klauschen, K.-R. M{\"u}ller, and W.~Samek.
\newblock On pixel-wise explanations for non-linear classifier decisions by
  layer-wise relevance propagation.
\newblock {\em PloS one}, 10(7):e0130140, 2015.

\bibitem{xai2}
D.~Baehrens, T.~Schroeter, S.~Harmeling, M.~Kawanabe, K.~Hansen, and K.-R.
  M{\~A}{\v{z}}ller.
\newblock How to explain individual classification decisions.
\newblock {\em Journal of Machine Learning Research}, 11(Jun):1803--1831, 2010.

\bibitem{chen2017attentive}
J.~Chen, H.~Zhang, X.~He, L.~Nie, W.~Liu, and T.-S. Chua.
\newblock Attentive collaborative filtering: Multimedia recommendation with
  item-and component-level attention.
\newblock In {\em Proceedings of the 40th International ACM SIGIR conference on
  Research and Development in Information Retrieval}, pages 335--344. ACM,
  2017.

\bibitem{deng2009imagenet}
J.~Deng, W.~Dong, R.~Socher, L.-J. Li, K.~Li, and L.~Fei-Fei.
\newblock Imagenet: A large-scale hierarchical image database.
\newblock In {\em Computer Vision and Pattern Recognition, 2009. CVPR 2009.
  IEEE Conference on}, pages 248--255. IEEE, 2009.

\bibitem{ex2}
J.~L. Herlocker, J.~A. Konstan, and J.~Riedl.
\newblock Explaining collaborative filtering recommendations.
\newblock In {\em Proceedings of the 2000 ACM conference on Computer supported
  cooperative work}, pages 241--250. ACM, 2000.

\bibitem{kingma2014adam}
D.~P. Kingma and J.~Ba.
\newblock Adam: A method for stochastic optimization.
\newblock {\em arXiv preprint arXiv:1412.6980}, 2014.

\bibitem{lapuschkin2016analyzing}
S.~Lapuschkin, A.~Binder, G.~Montavon, K.-R. Muller, and W.~Samek.
\newblock Analyzing classifiers: Fisher vectors and deep neural networks.
\newblock In {\em Proceedings of the IEEE Conference on Computer Vision and
  Pattern Recognition}, pages 2912--2920, 2016.

\bibitem{mcauley2015image}
J.~McAuley, C.~Targett, Q.~Shi, and A.~Van Den~Hengel.
\newblock Image-based recommendations on styles and substitutes.
\newblock In {\em Proceedings of the 38th International ACM SIGIR Conference on
  Research and Development in Information Retrieval}, pages 43--52. ACM, 2015.

\bibitem{xai1}
M.~T. Ribeiro, S.~Singh, and C.~Guestrin.
\newblock Why should i trust you?: Explaining the predictions of any
  classifier.
\newblock In {\em Proceedings of the 22nd ACM SIGKDD International Conference
  on Knowledge Discovery and Data Mining}, pages 1135--1144. ACM, 2016.

\bibitem{samek2017evaluating}
W.~Samek, A.~Binder, G.~Montavon, S.~Lapuschkin, and K.-R. M{\"u}ller.
\newblock Evaluating the visualization of what a deep neural network has
  learned.
\newblock {\em IEEE transactions on neural networks and learning systems},
  28(11):2660--2673, 2017.

\bibitem{xaideep}
W.~Samek, T.~Wiegand, and K.-R. M{\"u}ller.
\newblock Explainable artificial intelligence: Understanding, visualizing and
  interpreting deep learning models.
\newblock {\em arXiv preprint arXiv:1708.08296}, 2017.

\bibitem{shankar2017deep}
D.~Shankar, S.~Narumanchi, H.~Ananya, P.~Kompalli, and K.~Chaudhury.
\newblock Deep learning based large scale visual recommendation and search for
  e-commerce.
\newblock {\em arXiv preprint arXiv:1703.02344}, 2017.

\bibitem{ex3}
J.~Vig, S.~Sen, and J.~Riedl.
\newblock Tagsplanations: explaining recommendations using tags.
\newblock In {\em Proceedings of the 14th international conference on
  Intelligent user interfaces}, pages 47--56. ACM, 2009.

\bibitem{ex1}
Y.~Zhang, G.~Lai, M.~Zhang, Y.~Zhang, Y.~Liu, and S.~Ma.
\newblock Explicit factor models for explainable recommendation based on
  phrase-level sentiment analysis.
\newblock In {\em Proceedings of the 37th international ACM SIGIR conference on
  Research and development in information retrieval}, pages 83--92. ACM, 2014.

\bibitem{zhou2017deep}
G.~Zhou, C.~Song, X.~Zhu, X.~Ma, Y.~Yan, X.~Dai, H.~Zhu, J.~Jin, H.~Li, and
  K.~Gai.
\newblock Deep interest network for click-through rate prediction.
\newblock {\em arXiv preprint arXiv:1706.06978}, 2017.

\end{thebibliography}
\end{document}